\documentclass[journal]{sty/IEEEtran}

\usepackage{xcolor}

\usepackage{hyperref}

\usepackage{amsthm,amssymb,url}

\usepackage{graphicx}

\usepackage{multirow}
\usepackage{array}
\newcolumntype{P}[1]{>{\centering\arraybackslash}p{#1}}
\newcolumntype{M}[1]{>{\centering\arraybackslash}m{#1}}

\usepackage{amsmath}

\usepackage{caption}
\usepackage{subcaption}

\begin{document}
\title{Detecting broken Absorber Tubes in CSP plants using intelligent sampling and dual loss}

\author{M.A.~Pérez-Cutiño*,
        J.~Valverde,
        and~J.M~Díaz-Báñez% <-this % stops a space
\thanks{This work is partially supported by the European Union’s Horizon 2020 research
and innovation program under the
Marie Sklodowska-Curie grant agreement
734922, the Spanish Ministry of Economy and
Competitiveness (MTM2016-76272-
R AEI/FEDER,UE), the Spanish
Ministry of Science and Innovation
CIN/AEI/10.13039/501100011033/
(PID2020-114154RB-I00) and the European Union NextGenerationEU/PRTR (DIN2020-011317). \newline
The authors are with the Department of Applied Mathematics II, University of Seville, Seville 41092, Spain %
{\tt email: \{mpcutino, jvalverde, dbanez\}@us.es}}% <-this % stops a space
\thanks{M.A. Pérez-Cutiño and J. Valverde are with Virtualmechanics sl., Seville 41015, Spain.}% <-this % stops a space
\thanks{*Corresponding author: {\tt mpcutino@us.es}}
\thanks{Manuscript received -; revised -.}}

\maketitle

\begin{abstract}
Concentrated  solar power (CSP) is one of the growing technologies that is leading the process of changing from fossil fuels to renewable energies. The
sophistication and size of the systems require an increase in maintenance tasks to ensure reliability, availability, maintainability and safety. %In order to address failure detection, machine learning techniques have been typically considered in the industrial sector. In supervised learning, classification requires labelled data and the use of a large data set is mandatory.
Currently, automatic fault detection in CSP
 plants using Parabolic Trough Collector systems evidences two main drawbacks: 1) the devices in use needs to be manually placed near the receiver tube, 2) the Machine Learning-based solutions are not tested in real plants. We address both gaps by combining the data extracted with the use of an Unmaned Aerial Vehicle, and the data provided by sensors placed within 7 real plants. The resulting dataset is the first one of this type and can help to standardize research activities for the problem of fault detection in this type of plants. Our work proposes supervised machine-learning algorithms for detecting broken envelopes of the absorber tubes in CSP plants. The proposed solution takes the class imbalance problem into account, boosting the accuracy of the algorithms for the minority class without harming the overall performance of the models. For a Deep Residual Network, we solve an imbalance and a balance problem at the same time, which increases by 5\% the Recall of the minority class with no harm to the F1-score. Additionally, the Random Under Sampling technique boost the performance of traditional Machine Learning models, being the Histogram Gradient Boost Classifier the algorithm with the highest increase (3\%) in the F1-Score. To the best of our knowledge, this paper is the first providing an automated solution to this problem using data from operating plants.

%This work proposes supervised machine-learning algorithms for detecting broken envelopes of the Absorber Tubes in Parabolic Trough CSP plants. Moreover, we create a large data set, ATSet, obtained during real inspections and made it publicly available. To  the best  of our knowledge, ATSet is the first one of this type and would lead to a boost in research activities for this type of plants. %Our experiments validate the use of Dense-Sparse-Dense training for residual MultiLayer Perceptrons, and the learning capabilities of our system are demonstrated with the definition of a dual problem.

\end{abstract}

\begin{IEEEkeywords}
Concentrated solar power, Fault detection, Class imbalance, Machine Learning, Dual learning,  Dataset.
\end{IEEEkeywords}

% \fi
\IEEEpeerreviewmaketitle

% \section*{Multimedia Material}

% The Absorber Tubes dataset can be found in \url{https://atdataset.github.io/}.\footnote{ The full dataset will be made publicly available upon acceptance of the paper submission.}

\section{Introduction}

% ---------------------  JOURNALS to consider -----------------

% Reliability Engineering and System Safety
% Structural Health Monitoring

% Neural Computing and Applications
% Applied Intelligence
% International Journal of Energy Research
% Sustainable Energy, Grids and Networks
% IEEE Journal of Selected Topics in Applied Earth Observations and Remote Sensing

% some conferences
% https://2023.ieeesyscon.org/
% http://www.isdea.org/cfp.html
% https://ciss.jhu.edu/
% https://2023.refsq.org/

% transactions on geoscience and remote sensing
% pattern recognition
% transactions on signal processing
% transactions on sustainable energy

% ----------------------------------------------------------

\IEEEPARstart{B}{ack} in 1956, Marion King Hubbert proposed a concept related to the peak production of crude oil, which is expected to occur in early decades of the 21st century \cite{bardi2009peak}. If Hubbert theory is correct, then human kind is close to a disruptive change in their current way of living. Renewable energy resources arise as a promising alternative. Among the renewable energy resources, solar energy is, by far, the largest exploitable one, providing more energy in one hour than was consumed by humans in 2001 \cite{lewis2006powering}.
%In one hour, the sunlight strikes the Earth with more energy than all of the energy consumed by humans in an entire year \cite{lewis2006powering}.

\begin{figure}
    \centering
    \begin{subfigure}{0.49\columnwidth}
    \centering
        \includegraphics[width=\columnwidth, height=4cm]{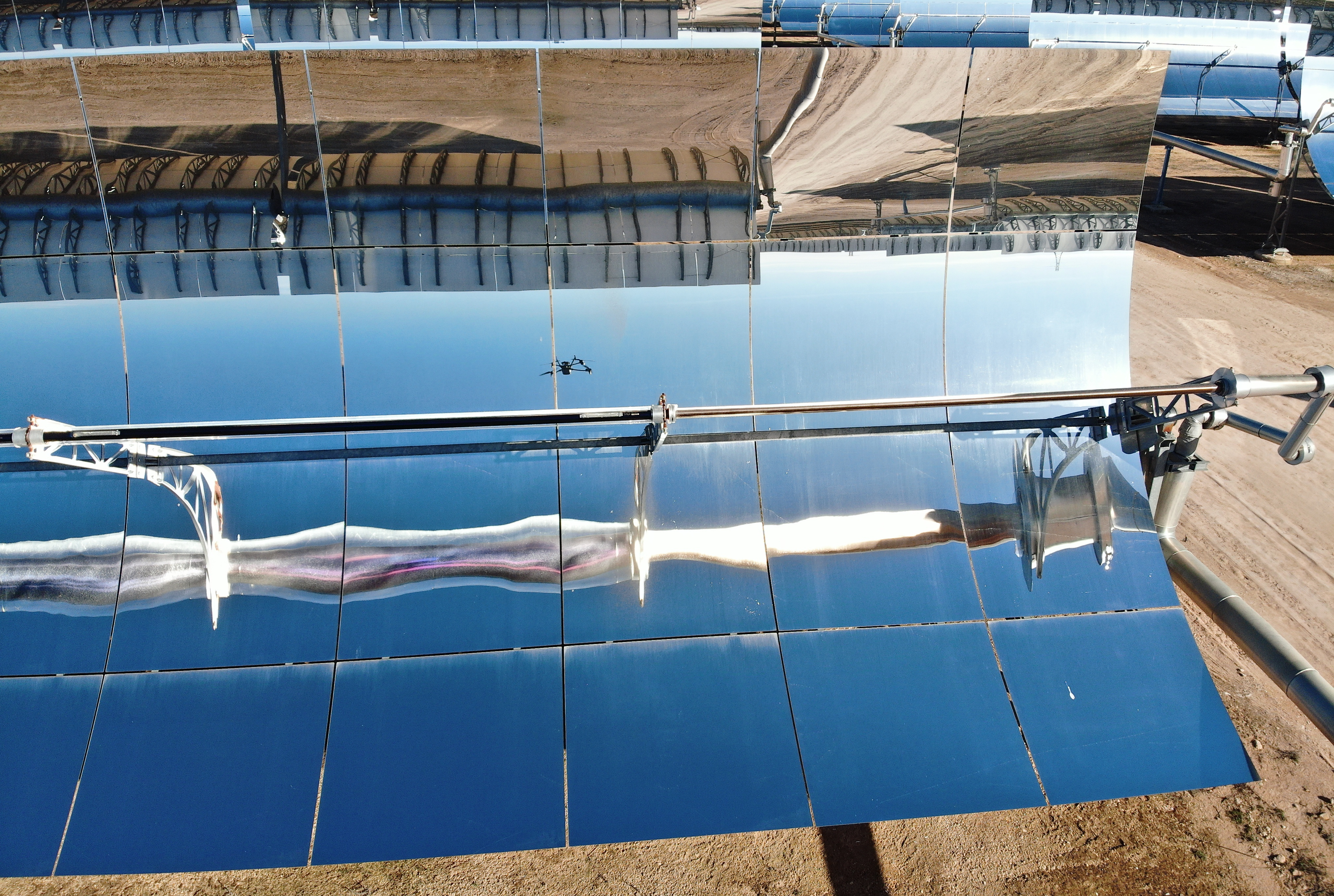}
        \caption{}
    \end{subfigure}
    \hfill
    \begin{subfigure}{0.49\columnwidth}
    \centering
      \includegraphics[width=\columnwidth, height=4cm]{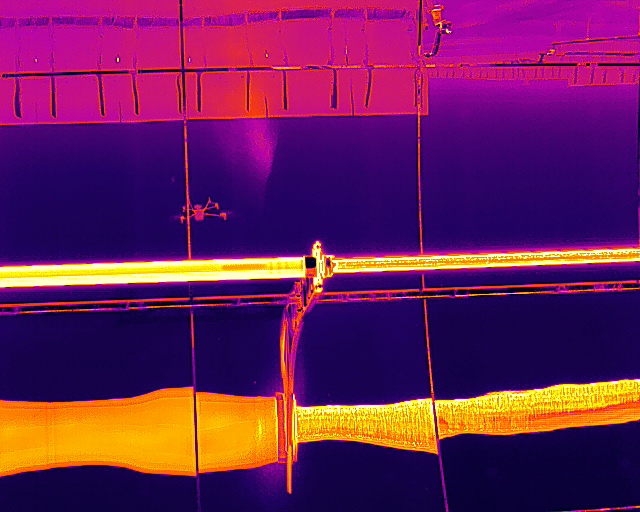}
      \caption{}
    \end{subfigure}
    \caption{Aerial image of a Parabolic Through Collector system captured by a drone with a traditional RGB camera (a), and an infrared camera (b). Images contain receivers with broken and non broken glass. Better viewed in color.}
    \label{fig:initial}
\end{figure}

Capturing and storing solar energy poses several challenges. Traditional  PV solar panels have an efficiency around 20\%, which implies that the majority of the energy coming from the sun rays is not captured. Concentrated Solar Power (CSP) plants represent a family of solar plants where the sun rays are concentrated over an specific surface, reaching efficiencies around 30\%. In this context, CSP plants using Parabolic Trough Collector (PTC) systems are based on parabolic-shaped mirrors that reflects solar radiation to an absorber tube located in the focal line of the parabola.  This system allows heating a transfer fluid to power a conventional Rankine thermal cycle. The absorber tube, also called receiver tube or heat collector element (HCE), is composed of a glass tube and a metal tube, with vacuum between them to reduce the heat loss. Even if the receiver tube is resistant to weather conditions, and has a good light transmittance performance, the glass envelope is prone to vacuum leakages which eventually turns into full glass breakage (see the right tube in Figure \ref{fig:initial})).

The detection of broken glass envelopes is crucial for keeping a correct functioning of the CSP plant and ensure reliability in the system. The heat loses of the HCE's without glass envelope increase almost 50\% due to activation of convective heat loses (vacuum loss), and by degradation of selective coating deposited onto the metal tube surface, whose aim is to control radiative heat loses. 
Fortunately, the number of broken envelopes in a productive CSP plant is relatively small, being around 1\% of the total number of absorber tubes. This is a significant difference between the prior probabilities of samples corresponding to broken and non-broken envelopes. From a Machine Learning perspective, this is known as the class imbalance problem \cite{japkowicz2002class}, which has a negative impact in the performance of classification algorithms. In addition, the minority class usually retains the highest interest from the learning perspective, and results in a higher cost when it is not well classified. A reliable system must account for this imbalance setting in order to be functional in real world environments.
%This significant difference between the prior probabilities of different classes in a dataset, known as the class imbalance problem \cite{japkowicz2002class}, has a negative impact in the performance of classification algorithms. Usually, the minority class retains the highest interest from the learning perspective, and results in a higher cost when it is not well classified. A reliable system must account for this imbalance setting in order to be functional in real world environments.

Techniques for inspecting CSP plants are grouped into liquid penetrant inspection (LPI), magnetic particle inspection (MPI), magnetic flux leakage (MFL), and visual inspection (VI). Of them, only VI can be used for inspecting the absorber tubes \cite{mesas2017development}. However, this task is inefficient when performed directly by humans as there are several kilometers of absorber tubes to inspect. % A possible approach is the use of Unmaned Aerial Vehicles (UAV) equipped with sensors to collect a variety of signals. The specific sensor and signal to use can be customized depending on the target application.
We present a solution based on the data gathered by an Unmaned Aerial Vehicles (UAV) inspecting several plants. The aerial platform capture pairs of RGB and thermal images, as in Figure \ref{fig:initial}. Thermal images are converted to grayscale to improve HCE's detection, and then a characteristic temperature value is associated to each receiver. Using the measured temperature values and operational data provided by the plants, the thermal conditions and the mechanical state of the HCEs are modeled to account for real heat loses on the solar field and mechanical safety, which %to 
avoid leakages or incidents. It is worth noting here that the glass envelope temperature provides indirectly the level of vacuum inside the annular chamber between steel and glass tubes. From the factory, a new HCE has a 10$^{-4}$ mbar pressure, almost vacuum which means no conduction or convection at all in the chamber. This desired vacuum can be jeopardized both by:
\begin{itemize}
    \item Ambient air entering the chamber due to cracks in both the glass or steel components, or even in the glass-to metal connection between both. This effect rapidly increases the pressure inside the vacuum chamber to ambient pressure ($\approx 1$ bar) of air, drastically increasing heat loses and therefore increasing the temperature of the glass envelope as seen from the termographic camera.
    \item H$_2$ filtering from the degraded HTF to the vacuum chamber though the stainless steel pipe due to the small size of the molecules (the smaller in nature and one with higher conductivity properties which increases conduction heat losses rapidly). Obviously, this effect was never envisioned by HCE manufacturers and plants developers and it is being tackled with different techniques and better maintenance and treatment of the HTF to avoid degradation . Monitoring the thermal state of the plant is one of the techniques at hand, as we are reporting on the present paper with the help of UAVs.
\end{itemize}
It is clear that the glass envelope temperature is the right indicator to detect loss vacuum by any of the effects explained above, which in turn end up with eventual glass envelope breakage and total loss of performance of the HCE. H$_2$ related vacuum loss is a slow process which can take months, even years to reach high pressure values and ruin the thermal performance of the HCE, therefore periodic surveying is mandatory to understand the real state of the solar field. On the other hand, air leakages by cracks is very fast and ends up with broken glass envelope very rapidly. In both cases, broken envelope is the final undesired state that we are willing to detect with the techniques proposed in the present work.

To the best of our knowledge, there is no dataset in the literature for the detection of broken glass in CSP plants. Publicly available data can improve the results of several research teams, and allows to standardize the evaluation of different methodologies over the same data input. We present the Absorber Tube Set (ATSet)\footnote{ATSet can be requested to the corresponding author for academic and research purposes.}, the first dataset for detecting broken-glass envelopes in CSP plants. We provide different benchmarks to our dataset using Machine Learning algorithms. %Additionally, we boost the performance of our models with the use of engineered features, different sampling techniques, and complex Deep Learning strategies.
Despite the high imbalance of the data, we increase the detection accuracy of the minority class with a negligible impact on the overall performance of the models, using engineered features, different sampling techniques, and complex Deep Learning strategies.

%Additionally, we propose a set of Machine Learning (ML) algorithms to tackle a classification problem: binary categorization of the glass envelope of HCEs. 
Our main contributions are summarized as follows:
\begin{itemize}
    \item We present ATSet, the first dataset for detecting broken-glass envelopes in CSP plants. A study of the impact of several variables related to productivity in CSP plants is conducted.
    \item We establish a benchmark for the broken-glass detection task, using several Machine Learning methods. We use domain knowledge to engineer new features, improving the baseline performance of our classifiers and mitigating the imbalance problem.
    \item We apply different sampling techniques to traditional Machine Learning methods such as Random Forest and Histogram Gradient Boost, increasing the performance of the algorithms.
    \item For the Artificial Neural Network, we define a dual problem to solve an unbalanced problem, and a balanced problem at the same time. Combining this strategy with Dense-Sparse-Dense training yields in the higher Recall for the minority class without harming the F1-Score.
    % \item {\color{blue}Rewrite...as a novelty...}We experimentally validate the improvements obtained by defining a dual problem, and using Dense-Sparse-Dense training for our Artificial Neural Network. This combination yields in the higher Recall for the minority class without harming the F1-Score.
\end{itemize}

The remainder of the paper is organized as follows: a literature review is conducted in Section \ref{sec:rw}; Section \ref{sec:problem_desc} formally describes our problem and the different models used; in Section \ref{sec:dataset} an analysis of the proposed dataset is performed; while in Section \ref{sec:ML} the ML algorithms and the DL techniques are properly defined. Sections \ref{sec:experiments} and \ref{sec:Ablation} present the experimental results of the proposed algorithms; and conclusions are summarized in Section \ref{sec:conclusion}.

\begin{figure*}
    \centering
    \begin{subfigure}{0.35\textwidth}
    \centering
        \includegraphics[width=\columnwidth, height=3.5cm]{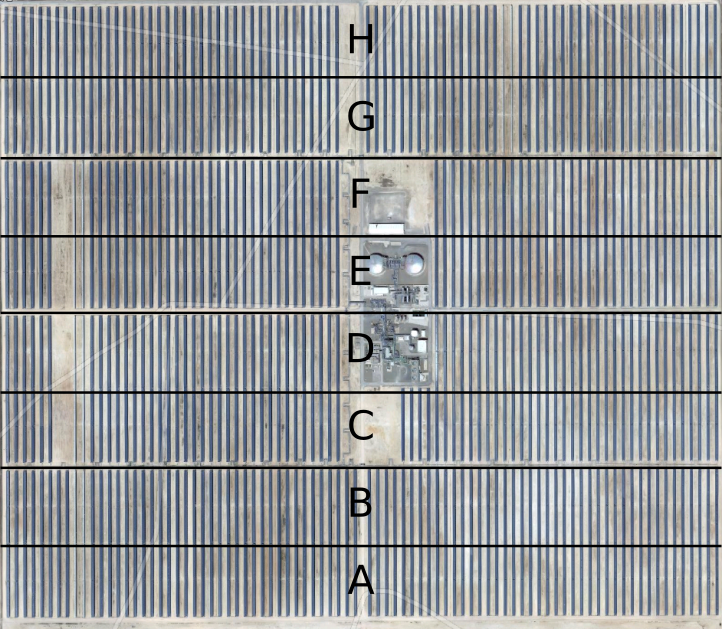}
        \caption{}
    \end{subfigure}
    \hfill
    \begin{subfigure}{0.27\textwidth}
    \centering
      \includegraphics[width=\columnwidth, , height=3.5cm]{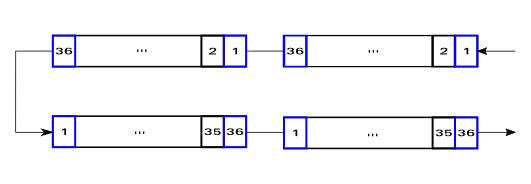}
      \caption{}
    \end{subfigure}
    \hfill
    \begin{subfigure}{0.35\textwidth}
    \centering
      \includegraphics[width=\columnwidth, height=3.5cm]{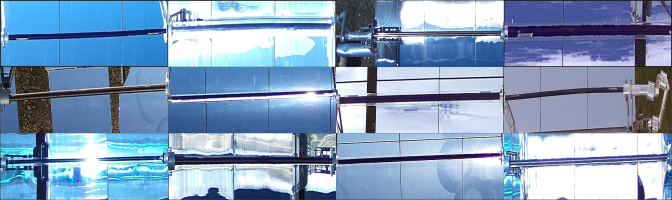}
      \caption{}
    \end{subfigure}
    \caption{Decomposition of a CSP plant (using PTC systems) by its fundamental parts. (a) geospatial image of a CSP plant, divided in 8 subfields, labeled from A to H. (b) loop of the HTF, passing through 144 HCEs divided in 4 stages. Blue rectangles define where the temperature of the HTF is measured by the plant. (c) subset of HCEs aerial images  captured by the UAV.}
    \label{fig:subfield_loop_tube}
\end{figure*}

\section{Related Work}
\label{sec:rw}

Machine Learning for safety applications have been an attractive field of research over the past years. 
%\cite{tang2019comparison,golcarenarenji2021machine,adhya2022performance}.
%The increase in the computational power and the large units of data storage, have turn the focus to more complex algorithms, such as Deep Learning. 
However, in the context of CSP plants, the use of these methods is an under-researched area. To provide better insights on different techniques, we divide the literature review into two parts: similar problems in photovoltaic systems, and similar problems in CSP plants. 
%For a general summary on Machine Learning methods for reliability and safety applications, see \cite{xu2021machine}.
For a general summary on Machine Learning methods for fault detection in industrial applications, see \cite{fernandes2022machine}.

% \cite{han2019field} A field-applicable health monitoring method for photovoltaic system  -->> Not ML

\subsection{Fault detection in photovoltaic systems using ML}

Decision Trees (DT) and Random Forest (RF) are popular methods applied to fault detection and diagnosis in photovoltaic (PV) systems \cite{benkercha2018fault,chen2018random,ziane2020detecting}. Different descriptive variables are defined, such as the ambient temperature, the irradiation and the operating voltage or the strings current of the PV array. Results obtained with RF evidence the superiority of this algorithm over DT \cite{chen2018random}. Other methods includes metaheuristics \cite{hazra2017efficient,das2018metaheuristic}; Naive Bayes \cite{niazi2019hotspot}; Support Vector Machines \cite{hafdaoui2022analyzing,jufri2019development}; and Gradient Boost Classifiers \cite{adhya2022performance}.

The use of Artificial Neural Networks establish an alternative to traditional classifiers. In \cite{aziz2020novel}, signals from the time domain are transformed to the frequency domain to generate scalograms that can be used as input of Convolutional Neural Networks (CNNs). The results of different classifiers where compared, being CNN and Random Forest the best ones when data was corrupted by noise. In the case study of \cite{chen2019deep}, Deep Residual Networks result in better performance than standard CNN when processing the data of current-voltage curve and the corresponding ambient conditions.

%PV systems differs from CSP plants in terms of design and functioning; therefore, similar problems can have totally different solutions. However, algorithms and data gathered for PV systems should be carefully studied, in order to apply custom transformations that successfully solve a given problem in CSP systems. To this end, we select Random Forest and Deep Residual Networks for fault detection in CSP plants using PTC systems. %For a comprehensive review of ANNs applied to fault detection in photovoltaic systems, see \cite{li2021application}.

\subsection{Fault detection in CSP plants}

Nowadays, CSP plants are considered as an emerging technology that could disrupt the energy production sector. Therefore, when compared to other types of solar power plants, the literature concerning automatic defects detection is scarce.

The works of \cite{espinosa2016vacuum,meligy2021iot} describe solutions for CSP problems without the use of Artificial Intelligence or massive UAV data collecting. In \cite{espinosa2016vacuum}, a methodology to obtain the gas composition 
%\textcolor{red}{, mainly the filtered H$_2$ from the HTF or air from ambient as described in the introduction} 
of the vacuum chamber of the receiver tube is presented. Using this information, it is possible to know if the vacuum has been lost, and measure the impact of the receiver's thermal performance. Their proposal yields on a non-destructive test with devices that needs to be manually operated and placed near the receiver tube. Therefore, they could only evaluate 1240 receivers, out of 13.248 composing the plant. On the other hand, \cite{meligy2021iot} proposes a distributed system for correcting the positioning of the mirrors. They use a set of accelerometer sensors to obtain the inclination angles, which can be used by a controller to align the mirrors and prevent long periods of efficiency loss. The testing environment was a Linear Fresnel Reflector (LFR), which is a CSP plant suitable for small-scale applications.

We are interested in an automatic solution for the problem of fault detection in CSP plants using PTC systems. This automation can be achieved by using ML algorithms capable of extracting useful information of the data collected in the solar plant in a massive and automated way with UAVs. The works of \cite{jimenez2017artificial,arcos2018concentrated,gomez2018cracks} are an initial attempt towards our target direction.

In \cite{jimenez2017artificial}, an artificial neural networks is used to determine the temperature of the pipe, using ultrasonic transducers. With this methodology it is possible to identify sudden changes in the temperature of the CSP plant, which can be related to faults such as corrosion. The data was collected in a test rig indoors. The work of \cite{arcos2018concentrated} proposes a Cloud Computing infrastructure to store the information collected by sensors in a group of CSP plants. Additionally, they studied two applications using ultrasonic waves as input of Artificial Neural Networks: detection of cracks, and prediction of the temperature in the absorber tube. The experiments were carried indoors with a pipe, by making a cut with six different depths. The accuracy for determining the temperature range in the test set was 80\%. Finally, \cite{gomez2018cracks} uses electromagnetic transducers to detect cracks and welds. They apply several signal processing techniques to find events that appears as peaks. These events can be related to cracks, edges, welds, or noise. The experimental setup is similar to those in \cite{jimenez2017artificial,arcos2018concentrated}; therefore, none of the above methods are tested in operating CSP plants, where signals can be disturbed by several factors.

% \cite{gonzalo2019review} is an extensive review that summarises the main degradation mechanisms and techniques used to detect, prevent, and mitigate failures.

The literature review for automatic fault detection in CSP plants using PTC systems evidences two main drawbacks of the proposed solutions: 1) devices needs to be manually placed near the receiver tube, 2) the ML solutions are not tested in real plants. We address both gaps by combining the data extracted with the use of a UAV, and the data provided by sensors placed within 7 real plants. We establish several benchmarks for the problem of detecting broken absorber tubes using ML algorithms. Additionally, as far as we know, this paper is the first in targeting the imbalance problem present in any functional CSP plant.

\section{Problem Description}
\label{sec:problem_desc}

CSP plants are traditionally structured in sets of subfields, as in Figure \ref{fig:subfield_loop_tube}(a). The number of HCEs may differ in each subfield; however, subfields are similarly organized as a set of sequential loops. In the context of CSP plants, a loop is defined as a four stage cycle, where the HTF enters in the first component and exits at the last, reaching a temperature jump of approximately 100$^\circ$C. Each component contains 36 HCEs, ordered according the path followed by the thermal fluid; see Figure \ref{fig:subfield_loop_tube}(b). At the beginning and the end of each stage, the temperature of the HTF is measured by physical sensors, which is used by the plant to correct the aiming point of the mirrors partial defocusing. When not carefully applied, these movements produce additional thermo-mechanical stress that can result in elastic and/or plastic bending, and eventually, in failure. If we add to the equation the normal ageing process that any industrial component is subject to, then we obtain some of the causes that results in the breakage of the glass envelope of the HCEs.

% Receiver that has lost their vacuum significantly increase their heat loss \cite{espinosa2016vacuum}. In this scenario, an extreme case can be related to the full loss of the glass envelope. Therefore, early detection of this issue  

Early detection of HCEs malfunctions is a key aspect in any productive CSP plant. However, manual inspection is a time consuming task, that has an additional difficulty related to the high temperatures of the solar field. On the other hand, Unmaned Aerial Vehicles are capable of inspecting a CSP plant in a few hours. A flying drone equipped with an infrared camera can collect thermal images with the information of the temperature of the glass surface of the HCE. Applying inverse engineering, it is possible to obtain the thermal conditions of the full receiver from the thermography combined with operational data (HTF temperature at the four control points of each collector) and weather conditions (ambient temperature, wind). Once the thermal state is obtained, where the contain of H$_2$ or air in the vacuum chamber is obtained as a direct consequence of the heat fluxes and temperatures measured, thermo-mechanical stress and deformation can be accounted for, and the safety of the component can be studied readily. This reverse engineering technology is owned by the company %proprietary to the company
Virtualmech\footnote{https://virtualmech.com/}, that has participated in the design of these components with some of the most important suppliers in the world.

The problem under study in this paper is the automatic detection of broken glass envelopes of absorber tubes in CSP plants, using numerical data. The envisioned solution has to be reliable in unbalanced scenarios, focusing the attention in the minority class.

\begin{figure}
    \centering
    \includegraphics[width=\columnwidth]{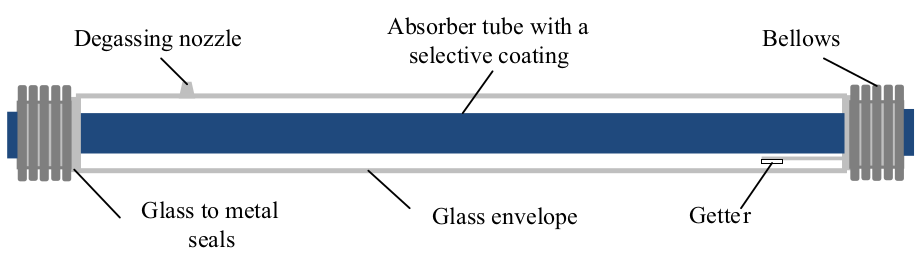}
    \caption{Components from a typical parabolic through receiver (extracted from \cite{espinosa2016vacuum}).}
    \label{fig:hce_schematic}
\end{figure}

The computational models developed are based on coupled FEM (Finite Element Method) and CFD (Computational Fluid Dynamics) simulation technology, in particular the Conjugate Heat Transfer (CHT) problem is solved at both fluid and solid domains. Therefore, the models include all the components of the HCE to be studied (see Figure \ref{fig:hce_schematic}), with their thermal and mechanical properties as a function of temperature. The purpose of these models is to obtain the resulting temperature and stress field in all the HCE components for the analyzed operating conditions. Two models are defined for each HCE:
\begin{itemize}
    \item Thermo-fluid coupled model: the transport of HTF in a turbulent regime is simulated with a mass flow rate of 6 kg/s inside the absorber tube for a given HTF temperature and the thermal constraints of the HCE in operation (concentrated radiation patterns, ambient conditions, etc.). From this model, the resulting temperature fields are obtained in all the components of the HCE, which are then exported to the mechanical problem. The internal radiation phenomena between each surface, and the conduction and convection losses have been modeled, depending on the vacuum pressure in the annulus space between glass envelope and metal tube. These losses are extracted following the theoretical model developed by Virtualmech, based on the NREL\footnote{National Renewable Laboratory US: \url{https://www.nrel.gov/}.} equations. In the external part of the HCE, radiation and convection phenomena to the environment have been simulated depending on the air temperature and presence/absence of wind.
    \item Mechanical model: the temperature fields of all the elements of the HCE are imported and 2 HCE's are simulated in series to faithfully represent the support conditions, and allow realistic thermal expansion along its axial axis. In addition to the above, the model is subjected to the action of gravity, internal HTF pressure of 40 bars and the weight of the internal HTF housed inside. From this model, the characteristic stress fields and strains of all the components of the HCE under study will be extracted for further safety analysis.
\end{itemize}
These models have been validated through several experiments in real CSP plants. Their output provides valuable information, such as the hydrogen pressure on the HCE annulus or the vacuum loss of the HCE, which are fundamental pieces of the proposed dataset. % At this point it is interesting to explain that the vacuum loss of the HCE is due to the permeabilization of hydrogen, coming from the degradation of the HTF, to the vacuum annulus through the stainless steel pipe carrying the fluid flow. The HTF degradation and hydrogen generation strongly depends of the fluid temperature, reaching high degradation velocities above 370-400$^\circ$C. 

% Despite of being a cumbersome task, the resolution of the proposed problem has great relevance not only to detect HCEs with broken envelopes, but also to correct the thermal model output. Specifically, when an HCE is known to be broken, the equations of the computational models can be re-evaluated to take into account the fact that the temperature has been measured over the pipe and not over the glass envelope. This translates into a feedback loop where the algorithms correct the output of the models, and then the models produce new data that can be used again by the algorithms. The process stops when there is no new broken prediction.

\section{Dataset}
\label{sec:dataset}

The lack of public available data describing the functioning of CSP plants prevents the research community from finding novel solutions to open problems. To overcome such difficulty, we build the Absorber Tubes Set (ATSet) with the results obtained during the inspections of 7 CSP plants. It contains the information of 155509 HCEs, involving 12 descriptive variables, and one target variable: the state of the glass envelope of the HCE. 

Table \ref{tab:data_vars} contains the information of the variables present in the ATSet, grouped by source. Virtual source refers to data extracted using the models described in the previous section. The temperature of the fluid of each HCE is obtained by interpolating the data provided by the plant in specific locations (the blue rectangles in Figure \ref{fig:subfield_loop_tube}(b)); therefore, it is grouped with the plant variables. On the other hand, the column Unit describe the variable by its measurement unit (like millibars, or watts per meter) or its type (like integer, or alphanumeric). Finally, the Range column establish the difference between each variable. For some Machine Learning methods, such as Artificial Neural Networks (ANN), the range value for each variable must be similar to ensure an effective learning \cite{sola1997importance}. Therefore, the data needs a normalization step before using ANNs. Additionally, a column for the graphical name is provided.

\begin{table*}
    \centering
    \begin{tabular}{|c|c|c|c|p{7cm}| c |}
    \hline
    \textbf{Variables} & \textbf{Range} & \textbf{Unit} & \textbf{Source} & \textbf{Description} & \textbf{Graphical name} \\
    \hline
    $T_g$  &  13$-$196 & ºC & Thermal Image  & \textit{Temperature} of the glass. & T\_HTF[C] \\
    \hline
    Loss  &  72$-$2068 & W/m & Virtual  & \textit{ Heat Loss} of the HCE. & Loss[W/m] \\
    Eff  &  0$-$1 & \% & Virtual  & \textit{Efficiency} of the HCE. & Eff \\
    $\textnormal{Eff}_r$  &  0$-$1 & \% & Virtual  & \textit{Efficiency} that the HCE should have for a given temperature of the fluid and ambient conditions. &  Eff\_ref \\
    PH2  &  $10^{\text{-3}}$$-$$10^3$ & mBar & Virtual  & \textit{Pressure} of the hydrogen.  & PH2[mBar] \\
    $T_{g-\epsilon}$ & 38$-$102 & integer & Virtual & The \textit{temperature} of the glass considering that the vacuum annulus contains air at 1Bar, \textit{minus a threshold}. & T\_glass Inf\_Limit \\
    $T_{g+\epsilon}$ & 58$-$122 & integer & Virtual & The \textit{temperature} of the glass considering that the vacuum annulus contains air at 1Bar, \textit{plus a threshold}. & T\_glass Sup\_Limit \\
    \hline
    $T_{HTF}$    &  269$-$411 & ºC & Interpolation &  \textit{Temperature} of the fluid. & T\_HTF[C] \\
    $\textnormal{HCE}_\textnormal{Loc}$  &  1$-$144 & integer & Plant & \textit{Location} of the HCE in the loop. & hce\_LocInLoop \\
    $\textnormal{HCE}_S$  &  A$-$H & character & Plant & Descriptor of the \textit{subfield} where the HCE is located. & hce\_number\\
    $\textnormal{HCE}_C$  &  1$-$94 & integer & Plant & Descriptor of the \textit{column} where the HCE is located. & hce\_column\\
    Plant & 0$-$6 & integer & Plant & Descriptor related to the name of \textit{the plant}. & plant\_name \\
    broken & 0$-$1 & binary & Plant & \textit{State} of the HCE: 0 when the glass envelope is not broken; 1 for the opposite case. & broken \\
    \hline
    \end{tabular}
    \caption{Dataset summary. Variables are grouped by source, showing its range and different measurement units. An additional column for description highlight in italics the more descriptive function of the variable. The target variable is binary, separating the data in two classes: 0 or Non-Broken, and 1 or Broken.}
    \label{tab:data_vars}
\end{table*}

The Virtual variables are computed assuming a glass envelope on the HCE. This assumption is enforced as there are not available methods for the automatic detection of broken glass envelopes. Therefore, the final values for the Virtual variables can be misleading for the broken samples\footnote{Broken samples refers to the samples of the data which class is Broken.} (recall that the temperature value is extracted from the thermal image). Our goal is to build an algorithm that properly classify the state of the HCE using this information. %It has to be noticed the complexity of the problem, as the algorithm needs to learn which data is associated to wrong values, i.e. which values are obtained with incorrect assumptions. 
As an additional complexity, class imbalance tips the balance in favor of non-broken samples: 151.833 samples of class 0, versus 3.676 samples of class 1. %\textcolor{magenta}{ An important remark needs to be made about Virtual variables. To prevent VirtualMech models from being reverse engineered, we add a small noise to corrupt the data. The noise is only applied over the non-broken samples, as they represent the majority of the data.}

\begin{figure}
    \centering
    \includegraphics[width=\columnwidth]{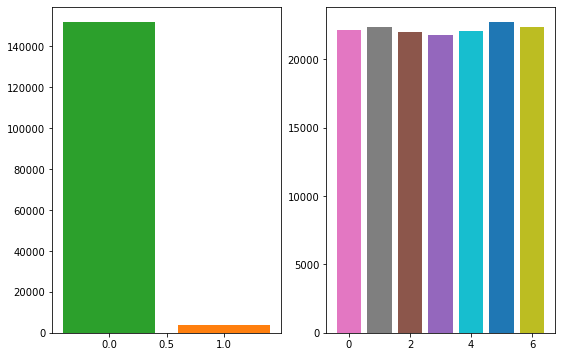}
    \caption{Data distribution for the dual problem. On the left, the data concerning to our main problem: broken pipes detection; on the right, the distribution of the data associated to the 7 plants in the surveys.}
    \label{fig:dual_balance_umbalance}
\end{figure}

Figure \ref{fig:dual_balance_umbalance} reflects the imbalance complexity associated to our problem. The left graphic represents the distribution of broken and non broken samples. On the other hand, the plant distribution is shown in the right graphic. In this case, all the plants are equally represented in the dataset. Therefore, we propose the use of the following dual problem:

\textit{Dual optimization problem (DOP):}
%by solving the following task:
%\textit{
given a set of features, classify the HCE by the state of its glass cover, and classify the HCE according to the plant where it is located. 

The first problem is our original problem, and the second one is the dual problem. 
%By defining a dual problem, our goal is to
The reason to use the
dual problem is to increase the performance of our algorithms in the minority class of the imbalance problem (the original), by solving at the same time a balanced problem (the dual). Our experiments will validate our proposal, specially when combined with other strategies at training time. 

\subsection{Analysis}

\begin{figure*}
    \centering
    \begin{subfigure}{0.24\textwidth}
    \centering
        \includegraphics[width=\columnwidth]{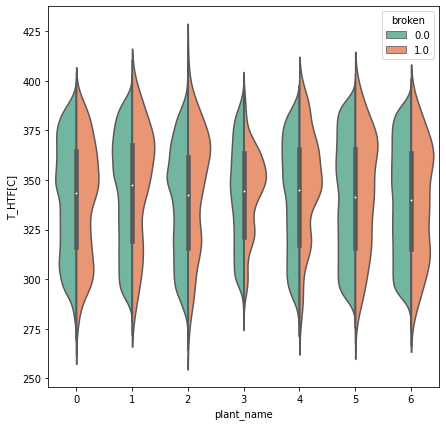}
        \caption{}
    \end{subfigure}
    \hfill
    \begin{subfigure}{0.24\textwidth}
    \centering
      \includegraphics[width=\columnwidth]{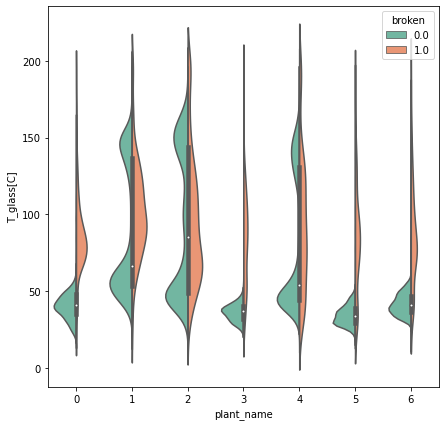}
      \caption{}
    \end{subfigure}
    \hfill
    \begin{subfigure}{0.24\textwidth}
    \centering
      \includegraphics[width=\columnwidth]{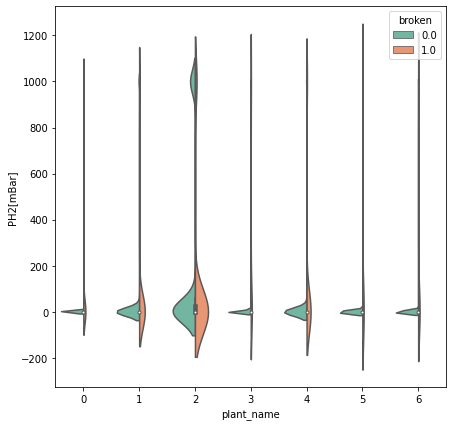}
      \caption{}
    \end{subfigure}
    \hfill
    \begin{subfigure}{0.24\textwidth}
    \centering
      \includegraphics[width=\columnwidth]{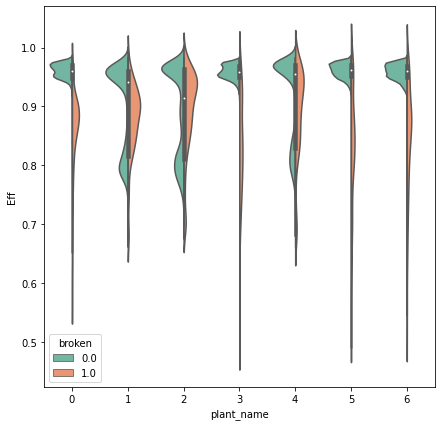}
      \caption{}
    \end{subfigure}
    \caption{Violin plots of four descriptors of the ATSet: (a) for the temperature of the fluid, (b) for the temperature of the glass, (c) for the hydrogen pressure, and (d) for the HCE efficiency. The efficiency plot shows the negative impact of the broken samples in the performance of the plants.}
    \label{fig:ots_violin_plots}
\end{figure*}

Figure \ref{fig:ots_violin_plots} shows a violin plot of some of the descriptive variables in the dataset. It must be noticed that our graphics does not take into account the number of samples on each class, only the probability density function of each variable across the different plants. The hills resulting from the density plot allows to understand how concentrated is the data around a value; however, one can not extract the minimum and maximum value of the variable directly from this plot. Another important aspect of violin plots is the visualization of the interquartile range, and the median value of the variable. All variables were analysed using violin plots, but only four of them were selected to study in this work due to its relevance in the functioning of the plant. 

The temperature of the fluid, see Figure \ref{fig:ots_violin_plots}(a), is around 340\textdegree C and has similar density function in all the plants, except for the broken samples of plant 3, that have a higher concentration of values around 350\textdegree C. The plants are designed to work with Thermal oil at temperatures from 300 to 400\textdegree C in the summer, and 250 to 350 in winter time, although in practice the maximum value is reduced from 400 to approximately 380 to prevent thermal oil degradation. % VER MAPC: However, the interquartile ranges shows that a high number of pipes transport the fluid at temperatures below 325\textdegree C, which translates into a possible non-optimized use of the plant resources. 

%On the other hand, 
The temperature of the glass envelope, see Figure \ref{fig:ots_violin_plots}(b), shows different density functions for the broken and non-broken samples, being the first between 60-120\textdegree C, and the later under 60\textdegree C. This is an expected outcome, as there is not glass covering the broken samples; therefore, the temperature is measured over the pipe. However, an unexpected result is that some non-broken samples has glass temperatures above 120\textdegree C. This can be produced by numerous factors, such as pipes bending, dirt over the glass surface,  among others. As result, it is not possible to define if the envelope of the HCE is broken only by measuring its temperature value.

The H$_2$ pressure in the vacuum chamber is another important factor of the correct functioning of the plant. In Figure \ref{fig:ots_violin_plots}(c), it can be observed the distribution of values between 0 and 100. The negative values are the result of the density function, and are not present in the dataset. However, the simulation results produce pressure values bigger than 100mBar; in such cases the result is fixed in 1000mBar, to establish a large separation from its neighbors\footnote{The pressure values are usually analysed after applying a logarithm of base 10, which is why the next power of 10 was selected as reference.}. It can be noticed that such high values are also present in non-broken samples. 

As the final variable to study, we select the efficiency, see Figure \ref{fig:ots_violin_plots}(d). For all the plants, it can be observed how a higher efficiency is related to non-broken samples, except from some particular cases where it drops to an $80\%$.

Violin plots highlights certain properties of the dataset. However, they also evidence that it is not possible to establish a clear separator between the target class if the descriptive variables are considered as individual components.

\subsection{Feature Engineering}

% Some of the features from Table \ref{tab:data_vars} are generated from the data set. The transformation of raw data into more descriptive variables is known as feature engineering. For instance, \textit{Delta\_Eff} is important for the plant to have a more informative value of the efficiency of the HCE.

The use of domain knowledge to simplify the learning process has been effective in several problems, such as time series analysis \cite{selvam2021tofee}, fraud detection  \cite{zhang2021hoba}, natural language processing \cite{lou2020automated}, among others. Inspired by this, we propose a set of new features to add into the dataset, both for training and evaluation. First, we normalize the glass temperature by subtracting \textit{$T_{g+\epsilon}$} with \textit{$T_g$}. Next, we transform the location values of the HCE. In the original dataset, the location in the loop is expressed as a number from 1 to 144. However, recall that each loop is divided in four stages of 36 HCEs. Each stage is subject to similar stress in terms of thermal requirements and movements. %of the figures (ball joints and double ball joints). 
Therefore, we change the location to be in the range of 1-36, using the remainder of the division between 144 and 36 (and set it to 36 in case is zero). Additionally, we add a new variable (\textit{structure\_in}) related to the presence of figures around the pipes. In this context, figures refer to components designed to hold the HCE in place and/or rotate it when the mirrors are out of focus. This is a binary variable. Finally, we transform the categorical variables into numerical codes.

\begin{figure}
    \centering
    \includegraphics[width=\columnwidth]{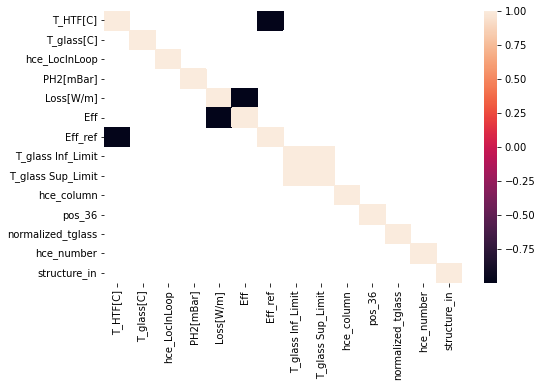}
    \caption{Subset of the correlation matrix of all the variables. Colored cell corresponds to variables highly correlated; i.e., with a correlation value bigger than $0.98$, or smaller than $-0.98$.}
    \label{fig:correlation}
\end{figure}

A correlation analysis is performed over the full dataset with the new features added. The goal is to discard variables with a linear relationship, identified by high (near to 1) or small (near to -1) correlation values. As it can be observed in Figure \ref{fig:correlation}, the group of variables with similar trends are: 1) \textit{Loss} and \textit{Eff},  which is understandable as the loss and the efficiency of the HCe's provide a similar information; 2) \textit{$T_{g+\epsilon}$}, \textit{$T_{g-\epsilon}$} have a correlation value of 1 since they only differ by a constant threshold; 3) \textit{$\textnormal{Eff}_r$}, \textit{$T_{HTF}$}, which can be explained as follows: $T_{HTF}$ determines the heat source from which the losses to the ambient are decremented for a given HCE. $\textnormal{Eff}_r$ is just the calculation of performance of this HCE at reference conditions ($\textnormal{Eff}_r = \frac{Q_f}{Q_i}$, where $Q_i$ is the heat introduced in the system and $Q_f$ = $Q_i - Q_l$ is the heat effectively transferred to the fluid, with $Q_l$ representing heat losses to ambient). Therefore, $\textnormal{Eff}_r$ has a linear inverse relationship to the temperature of the fluid (the bigger $T_{HTF}$ the bigger $Q_l$ and smaller $\textnormal{Eff}_r$) and a linear relation to the heat loss. However, for the threshold established, only one of these relationships is observable in the correlation matrix. %[\textcolor{red}{The reference efficiency is directly related to the HTF temperature as it determines the heat source from which the loses to ambient are decremented. It is more correlated to HTF temperature than the heat losses due to the fact that reference efficiency is adimesionalized with the energy input on the system while the heat losses are not, and therefore depend more on the specific ambient conditions.}].

The final descriptive variables in the featured engineered set are: \textit{$T_{HTF}$, $T_g$, $\textnormal{HCE}_\textnormal{Loc}$, PH2, Loss, $\textnormal{Eff}_r$, $T_{g+\epsilon}$, pos\_36, normalized\_tglass, $\textnormal{HCE}_C$, $\textnormal{HCE}_N$} (the numerical code of \textit{$\textnormal{HCE}_S$}), and \textit{structure\_in}.

\section{Machine Learning Algorithms}
\label{sec:ML}

\subsection{Deep Residual Network}

The baseline architecture for the neural network used is similar to the one proposed in \cite{perez2022ornithopter}. This model has been effectively tested on unbalanced datasets with more than 10 classes. It is composed of several linear layers with residual connections \cite{he2016deep}, layer normalization \cite{ba2016layer}, and dropout \cite{hinton2012improving} for regularization. The model is a Deep Residual Network (DRN) with 777 792 neurons.

The loss function used for this model is:
\begin{equation}
    L_g = -\sum_i w_ip_i(x)\log q_i(x),
    \label{eq:categorical_cross_entropy}
\end{equation}
which defines a weighted cross entropy to consider the different frequencies of the classes adopted. In the above formula, $p_i(x)$ defines the probability of sample $x$ of belonging to class $i$, $q_i(x)$ defines the probability predicted by the network, and $w_i$ is the weight value considered for class $i$. As our problem is defined for two classes, $L_g$ can be expressed as:
\begin{equation}
    L_g = -w_0p_0(x)\log q_0(x) - w_1(1-p_0(x))\log q_1(x).
    \label{eq:BCE}
\end{equation}
To produce small loss values for the majority class, we consider the percentage of samples of class $i$, $0\leq P_i\leq 1$, in the computation of the weight value. We set:
\begin{equation}
w_i = (1-P_i)^\alpha,
\end{equation}
where $\alpha$ is a parameter designed to control the class relevance. Its impact is studied on Section \ref{sec:Ablation}.

We add complex Deep Learning strategies for boosting the performance of our baseline model. We describe dual loss and Dense-Sparse-Dense (DSD) training in the following subsections.

\subsubsection{Dual Loss}

We propose to extend $L_g$ loss function by taking advantage of duality optimization. Several works, as for instance \cite{zhu2020deep,nagi2021ruf}, evidence the improvements obtained with this strategy in terms of performance of the models. We define a new problem where, given the data of the HCE, the goal is to predict in which plant the HCE is located. Solving this specific task has no interest to our original problem, but it can help the learning process because a subset of the weights of the DRN is shared across the two classification branches. % Additionally, the dual problem is well balanced, which can translate into more accurate results of our models.

It is worth mentioning the differences between using this approach and the method where the plant value is used as input of the network. In the second scenario, the model can not scale to plants that are not present in the training data. On the other hand, our approach is independent from the plant where the HCE is located, even if the model gives a possible location for the HCE. For data coming from new plants, the location output can be interpreted as the plant from the training data with more similarities to the new plant.

We define a new loss $L_p$ for the this task. The loss value has the same formulation as Equation \ref{eq:categorical_cross_entropy}. The dataset contains information of 7 different plants, and each one of them provides a similar amount of samples. Therefore, we  set $w_i=1$ for all classes considered in $L_p$. Finally, the dual loss is defined as:
\begin{equation}
    L_{dual} = \beta_1*L_g + \beta_2*L_p,
\end{equation}
where $(\beta_1, \beta_2)$ represent the contribution of each individual loss.

\subsubsection{DSD Training}

Dense-Sparse-Dense training is a regularization technique designed for improving the performance of the optimizer \cite{han2016dsd}. This algorithm decompose the training phase into three stages: a Dense (D) training for adjusting the weights of the network and learn their importance; a Sparse (S) training where the unimportant connections are pruned (depending on the sparsity parameter), and only the most relevant weights are updated; and a final Dense (D) phase where the model is retrained with all its parameters.

The application of this technique only affects the training time. In addition, it only uses an extra argument if compared to standard training: the sparsity $s$ of the model in the second phase, i.e. the percent of important connections that are going to be trained. In this context, the important connections are updated in each layer of the network in the following manner: search for the $1-s$ percent of connections associated to the lower values in the tensor of weights, and freeze them. By freezing this connections, only the important edges are updated during back-propagation in the Sparse phase. Note that all connections are used during the forward phase. % DSD training have been vastly used during the learning phase of Convolutional Neural Networks (CNN) \cite{georgescu2019local,tian2020meta} and Recurrent Neural Networks (RNN) \cite{han2016dsd}. However, to the best of our knowledge, this is not the case of Multilayer Perceptrons with residual connections.

The Adam optimizer was used during training. The parameters for the standard training phase were the ones recommended in \cite{kingma2014adam}. For the Dense-Sparse-Dense training, we change the learning rate (\textit{lr}) in each phase: first Dense, \textit{lr}=0.01; Sparse, \textit{lr}=0.001; and second Dense, \textit{lr}=0.0001. The idea is to reduce the weight update value in phases were the network is more confident about the solution.

\subsection{Random Forest}

Random Forest is a popular supervised Machine Learning method. %, successfully applied to remote sensing data \cite{zhu2021flood}, and simulated data \cite{perez2022ornithopter}. 
It is supported by a set of Decision Trees, where each element is trained over random samples of the training set, usually with a subset of the available features. This technique endow different learning capabilities to each tree in the forest, without overfitting when more trees are added to the system \cite{breiman2001random}.  

To measure the importance of each attribute before creating a new node in the tree, the Gini Index is used. Its value is obtained by deducting the squared probabilities $P_j$ of each class from 1:
\begin{equation}
    GI = 1 - \sum_j (P_j)^2.
\end{equation}

The architecture of the Random Forest is obtained with a Random Grid Search over a set of possible parameters, depicted in Table \ref{tab:rf_params}. The final results are described in Section \ref{sec:experiments}.

\begin{table}
    \centering
    \begin{tabular}{|c|M{2.5cm}|M{2.9cm}|}
    \hline
    \textbf{\# estimators} & [50, 70, 90, ..., 1000) &  Numbers of trees in the forest. \\
    \hline
    \textbf{max features}  & [sqrt, half, all] &  Maximum number of features considered at each split. \\
    \hline
    \textbf{max depth}    & [10, 20, ..., 100) & Maximum depth of the trees. \\
    \hline
    \textbf{class weight}  &  [balanced, equal, \{0: 0.2, 1: 0.5\}] & Weight associated with classes. \\
    \hline
    \textbf{min samples split} & [2, 5, 10] & Minimum number of samples required to split a node. \\
    \hline
    \textbf{min samples leaf} & [1, 2, 4] & Minimum number of samples required at each leaf node. \\
    \hline
    \end{tabular}
    \caption{Grid of parameters considered to build the Random Forest.}
    \label{tab:rf_params}
\end{table}

\subsection{Histogram Gradient Boost Classifier}

Gradient Boost Classifiers are ensemble algorithms based on additive models. Like in RF, we use a set of Decision Trees as baseline models. However, the learning process for each tree is simplified by learning the residual/miss-predictions of the previous tree. % Popular in industry \cite{zhang2022anomaly,liu2022multi}.
For big datasets with continuous variables, the features can be organized into bins to accelerate the training process. In such cases, the algorithm is known as Histogram Gradient Boost Classifier (HGBC).

The proposed HGBC is based in \cite{ke2017lightgbm}. The maximum number of beans for each variable is set to 255. HGBC is trained using a binary cross entropy with a learning rate of $0.1$.

\subsection{Re-sampling techniques}

Re-sampling techniques is a common approach for solving practical problems in industry. %\cite{dodangeh2020integrated,barzegar2021improving}.
There are several strategies which are usually categorized as over-sampling (OS) or under-sampling (US) techniques.

Over-sampling algorithms are used to increase the number of samples of a given class. There are different techniques that can be applied, like generating virtual samples or simply cloning the samples of a given class. For generating virtual samples we tried SMOTE \cite{chawla2002smote}, but it did not improve our baseline results. Therefore, we select the minority class and simply replicate its samples in the training set; then we use a RF to learn from this new set. RF takes a portion of the samples of the dataset for training new Decision Trees. Therefore, this combination is selected to boost the performance of each individual tree.

Under-sampling techniques can be used with two objectives: one the one hand, to select the best samples of the dataset to reduce the training time; on the other hand, to decrease the number of samples of the majority class in the training set to obtain better generalization. For the latter, we select Random Under-Sampling (RUS) as it results in a higher increase in performance when compared to other techniques. This algorithm randomly discards samples of the majority class before each step of the training loop. This strategy was successfully applied both to RF and a classifier composed of 50 HGBC.

% For selecting the best samples of the dataset, we apply uncertainty sampling. This method selects the samples near to the decision boundaries of a given classifier. This samples are assumed to be the hardest to classify. Therefore, a classifier with a good performance over the uncertainty set can provide a good generalization. For reducing the number of samples of the majority class, we apply Random Under-Sampling (RUS) to RF and a classifier composed of 50 HGBC.

\section{Experiments}
\label{sec:experiments}

We perform a set of experiments to train and evaluate the proposed dataset.

\begin{itemize}
    \item \textit{Feature Test}: The original variables of the dataset are compared against the feature engineered variables, by visualizing the output of a T-distributed Stochastic Neighbor Embedding(t-SNE) \cite{van2008visualizing} with two dimensions. The goal is to observe the benefits from adding the new features.
    \item \textit{Balance Test}: A fixed classifier is trained and tested with datasets matching an imbalance percent of $p$. The behaviour on each scenario is studied.
    \item \textit{Standard Test}: RF, DRN, and HGBC are trained over the proposed training set to predict the broken pipes. Evaluation is obtained over the proposed testing set.
    \item \textit{Advanced Test}: We study the behaviour of our classifiers using different sampling techniques. In addition, we present the benefits of training the DRN with a dual loss and a DSD strategy.
\end{itemize}

The metrics considered to evaluate our proposal are:
\begin{equation}
    Precision = \frac{TP}{TP + FP},
\end{equation}
\begin{equation}
    Recall = \frac{TP}{TP + FN},
\end{equation}
\begin{equation}
    F1 = \frac{2* Precision * Recall}{Precision + Recall},
\end{equation}
where TP, FP, FN stand for True Positives, False Positives and False Negatives, respectively. In addition, we compute the macro average F1-score (M-AVG F1). This metric does not take into account the class imbalance; therefore, it gives the same importance to each class when computing the average. Finally, 80\% of the dataset was used for training, and the remaining 20\% for evaluation. The results described are obtained for the evaluation (test) set.

\subsection{Feature Test}

The use of domain knowledge to obtain more descriptive features provides better insights about the target problem. We support this intuition by applying t-SNE on the original training set (OTS), and the training set with engineered features (FETS). t-SNE is an unsupervised algorithm with focus on keeping similar data points closer on a lower-dimensional space. Outliers has a low impact on t-SNE, and it is only used for visualization. Visualizing the features allows to assess the separability of the data.

\begin{figure}
    \centering
    \begin{subfigure}{0.49\columnwidth}
    \centering
        \includegraphics[width=\columnwidth, height=4cm]{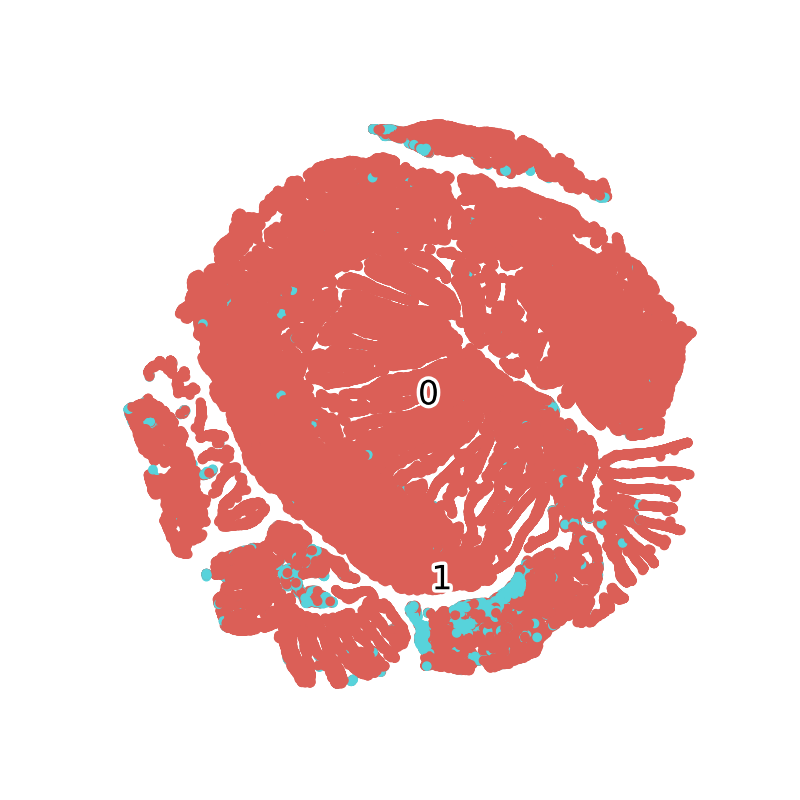}
        \caption{}
    \end{subfigure}
    \hfill
    \begin{subfigure}{0.49\columnwidth}
    \centering
      \includegraphics[width=\columnwidth, height=4cm]{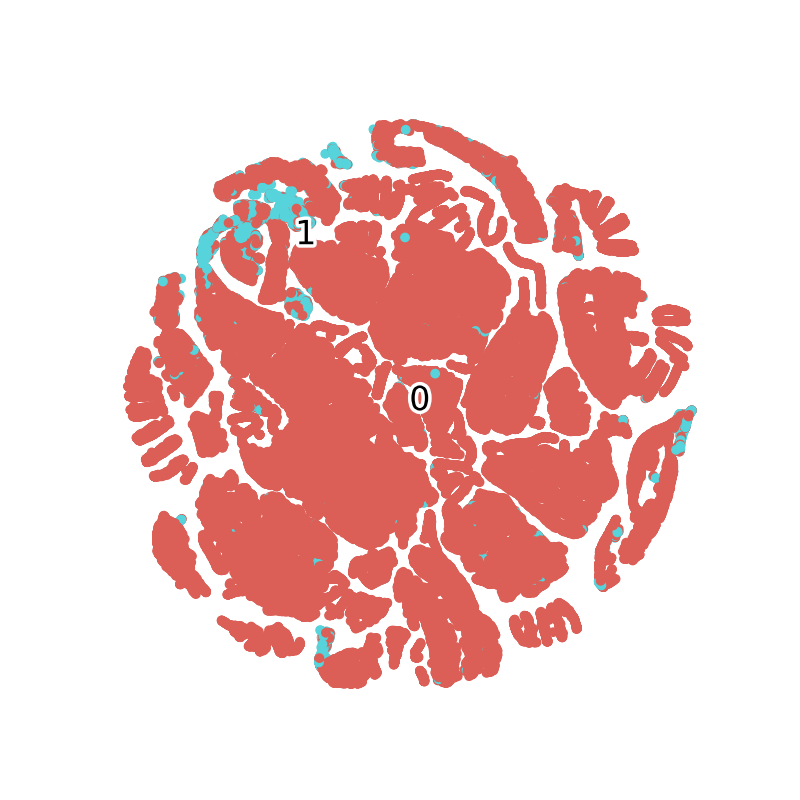}
      \caption{}
    \end{subfigure}
    \caption{t-SNE applied on OTS (a) and FETS (b). Class 0 has an elevated predominance, but it is more isolated on (b).}
    \label{fig:tSNE_broken_glass}
\end{figure}

We apply t-SNE to obtain embedding of two dimension, using a perplexity of 60, and 1000 iterations. The algorithm is applied both for OTS and FETS. Figure \ref{fig:tSNE_broken_glass} shown the results of coloring the output of t-SNE using the broken glass information in the dataset. It can be noticed the high imbalance on the dataset, in addition to the difficulties of properly separating each class. However we can observe more islands in Figure \ref{fig:tSNE_broken_glass}(b), which corresponds to the feature engineered data. Class numbers are placed using the coordinates of the mean value of all points corresponding to the same class.

\begin{figure}
    \centering
    \begin{subfigure}{\columnwidth}
    \centering
        \includegraphics[width=\columnwidth]{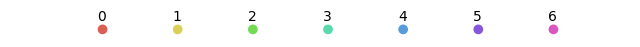}
    \end{subfigure}
    \begin{subfigure}{0.49\columnwidth}
    \centering
        \includegraphics[width=\columnwidth, height=4cm]{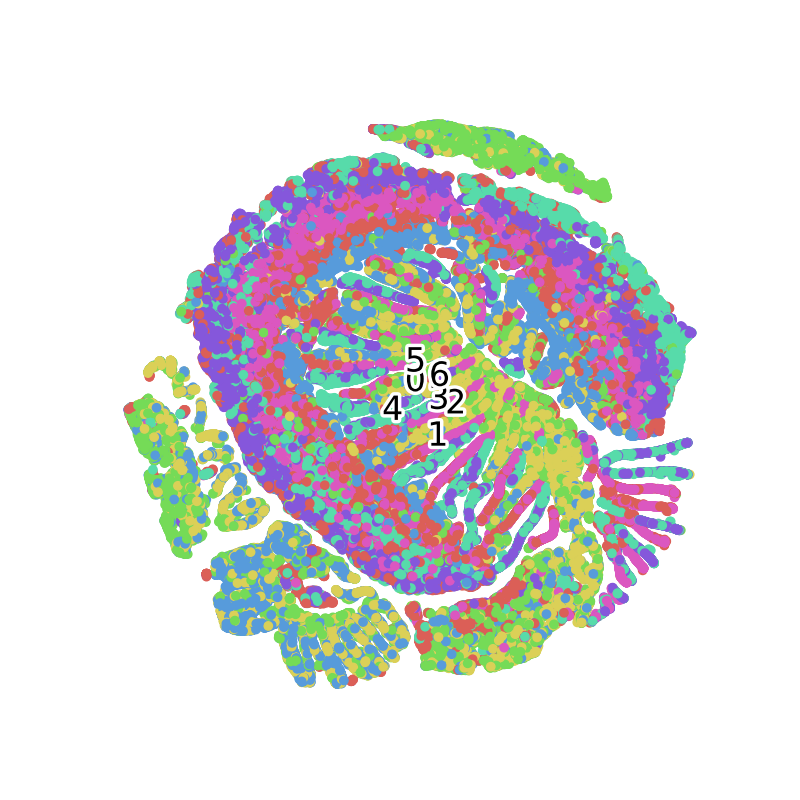}
        \caption{}
    \end{subfigure}
    \hfill
    \begin{subfigure}{0.49\columnwidth}
    \centering
      \includegraphics[width=\columnwidth, height=4cm]{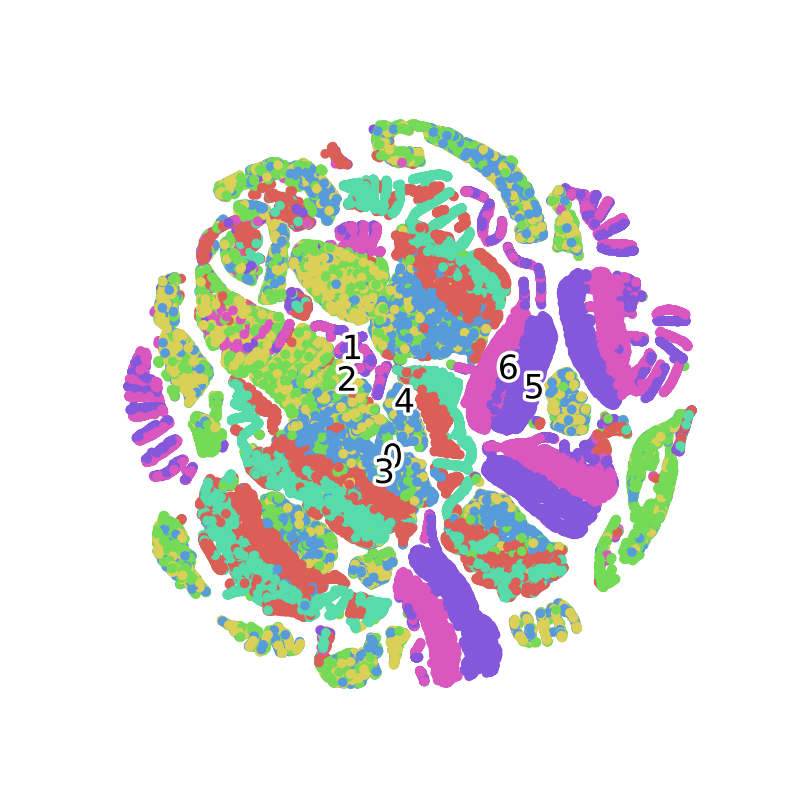}
      \caption{}
    \end{subfigure}
    \caption{t-SNE applied on OTS (a) and FETS (b), for the dual problem. Class represented are the different plants present on the dataset (best viewed in color). Separation is clearer for FETS, and the mean value per class is more disperse.}
    \label{fig:tSNE_plant_name}
\end{figure}

Figure \ref{fig:tSNE_plant_name} depicted a similar result when using the plant value for coloring the data points resulting from t-SNE. The mean value for each class is similar when using OTS. On the other hand, FETS derives on a better separation between points of different classes. These results support the intuition of using FETS as input for our classifiers, specially for the dual problem.

\subsection{Balance Test}

The goal of the test is twofold: on the one hand, it showcases the learning capabilities of the classifiers when the imbalance of the dataset is given by $p$; on the other hand, it evidences the complexity of our problem. We perform the balance test by randomly splitting the training set into a reduced training set and a reduced test set. Both the reduced training set and the reduced test set contains a proportion of broken pipes that match the percent value $p$. As the initial proportion is around 2 percent, we consider the integers in the interval [2, 50] as possible values for $p$.

Forcing $p$ as the proportion of broken pipes samples in the reduced sets implies that some of the non broken samples have to be discarded. In order to maintain a representative set of the data, the reduction process consider the positioning per plant for the non broken data. Therefore, if a plant contains $b$ broken samples, then $n$ non-broken samples in each possible location (from 1 to 36) are randomly selected from the rest of the data of the plant. $n$ is given by the formula:
\begin{equation}
    n = \frac{\frac{b}{p'} - b}{36} = \frac{b(1-p')}{36p'},
\end{equation}
where $p'=p/100$.

\begin{figure}
    \centering
    \includegraphics[width=0.8\columnwidth]{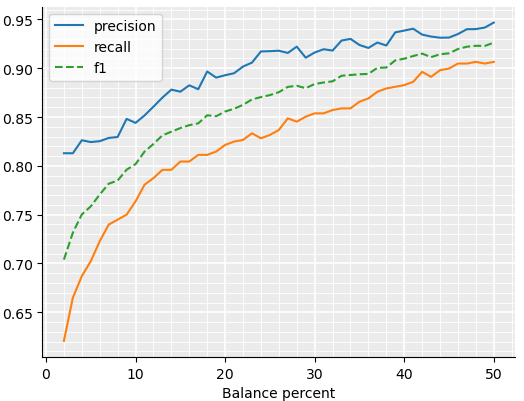}
    \caption{Balance test for a fixed Random Forest. More balance in data derives in a better performance. Training set was spitted in two: one set to train and other to validate. Metrics are computed over the broken pipes samples of the validation set.}
    \label{fig:balance_test_rf}
\end{figure}

We fix a Random Forest with 100 estimators, and none restriction on the depth of the trees. Gini is selected as the function to measure the quality of the split, and the weights associated to each class $c$ are adjusted to the inverse of the frequency of $c$ in the training set. Figure \ref{fig:balance_test_rf} depicted the performance of this classifier when the value of $p$ changes. For the highest imbalance, the Recall of the classifier is around $60\%$; when the data is balanced, this metric gets closer to $90\%$. However, we are interested in \textit{real world} scenarios, where the class imbalance is present. Detecting broken glass envelopes is the main interest of this work. Therefore, our goal is to increase the Recall value for this class, while preserving the F1-score. In this manner, our algorithms will be able of identify the broken envelopes while keeping a small rate of false positives.

\subsection{Standard Test}

In this experiment, we obtain the results of the proposed classifiers in the full test set. The classifiers are evaluated differently. In the case of Random Forest, a grid of possible parameters is defined. We randomly select 50 configurations and train the resulting RF using k-fold with $k=3$. Results are presented for the classifier with the highest F1-score in the cross-validation\footnote{Training data is randomly split in a training and validation set} process. A similar approach was followed for computing the parameters of the HGBC. In this case, we only modify the learning rate parameter. On the other hand, DRN is tested by changing the value of $\alpha$, using $L_g$ as loss function. Training set is randomly divided in a reduced training set and a validation set, and we do not perform cross validation for this experiment. Therefore, the training-validation split is preserved for different values of $\alpha$.

Results from the standard test are presented in Table \ref{tab:tab_standard_test}. If compared to the previously fixed RF, the random search derived in a RF with 800 estimators, and a maximum depth of 80 for every tree. In addition, the class weight considered was 0.2 for class 0, and 0.5 for class 1; the minimum number of samples to split a node was 5; the maximum number of features considered at each split was the square root of the total number of features; and the minimum number of samples required at each leaf node was 2. The result observed in Table \ref{tab:tab_standard_test} is consistent with the proposed goal: obtain a RF with a similar F1-score, but with a higher recall. If compared to the values obtained for the version of Figure \ref{fig:balance_test_rf} with more imbalance, the Recall has increased around a 10\%. For DRN, we select $\alpha=0.5$, as it results in the best trade-off between Recall and Precision. Compared to RF, the Recall increases by 8\%, maintaining a similar F1-score. Finally, HGBC was trained over 1500 iterations with a binary cross entropy loss, and a learning rate of 0.05. It has a similar F1-score than RF, but the Recall of the minority class is inferior.

\begin{table}
    \centering
    \begin{tabular}{|c|c|c|c|c|}
    \hline
    Algorithm & Precision & Recall & F1 Score & M-AVG F1 \\
    \hline
    DRN     &  63 & 78 & 70 & 84  \\
    RF      &  73 & 70 & 72 & 86 \\
    HGBC    &  78 & 66 & 72 & 86 \\
    \hline
    \end{tabular}
    \caption{Precision, Recall and F1 Score of the proposed classifiers for the broken samples of the test set. M-AVG F1 stands for the macro average of the F1 score, and it is computed for the entire test set.}
    \label{tab:tab_standard_test}
\end{table}

\subsection{Advanced Test}

% \subsubsection{Uncertainty sampling}

% We try to find the $u$ samples nearest to the decision boundaries using uncertainty sampling. As measure of uncertainty, we select the samples where the RF probability prediction is closer to 0.5 (assuming 1 as maximum). Figure \ref{fig:uncertainty_sampling_rf} depicted the results obtained for the testing set when the value of $u$ changes according to the considered balance percents. It is important to notice that when the training set is more balanced the value of $u$ decreases, because the classifier is more confident about its predictions. Additionally, the imbalance of the testing set does not change, which implies that the generalization is harder for lower values of $u$. 

% In this particular test, all metrics were stable across different balance percents up to 32\%; in this point, precision drops drastically. This is an interesting result as it evidences the impact of data selection: with better samples the performance can be maintained while reducing the training time. 

% \begin{figure}
%     \centering
%     \includegraphics[width=0.95\columnwidth]{figs/uncertainty_sampling_rf.png}
%     \caption{Results using uncertainty and distribution sampling over the training set. Values of metrics are obtained by changing the proportion of broken and non broken data in the training set. The test set does not change; therefore it maintains the original imbalance.}
%     \label{fig:uncertainty_sampling_rf}
% \end{figure}

\subsubsection{Ensemble of RF}
We propose a new test with Random Forest, where a voting scheme is evaluated for an over-sampling technique. The idea is to divide the training set in $t$ datasets such that: 1) all broken samples are present on any subset, 2) broken data represents a $p$ percent of the total data in any subset, 3) the union of the non broken samples of the subsets match the non broken samples in the training set. We select $t=8$, which derives in $p\approx 17$. A total of $t$ Random Forests are randomly initialized and trained over the subsets. Finally, the predictions are obtained by a majority voting process, i.e., the class where most of the classifiers agree is the output of the algorithm. Class 0 is preferred in case of ties, as it has a higher frequency in the dataset. We refer to this classifier as V-RF, for commodity.

\subsubsection{Random Under Sampling}

We apply random under sampling for the majority class in the training set. This technique is integrated into the Random Forest classifier, and a set of 50 Histogram Gradient Boost classifiers. The set of HGBC is evaluated using a voting scheme. We refer to this technique as Balanced HGBC, or B-HGBC, for commodity. We add the tag RUS when presenting the results for these classifiers.

\subsubsection{Dual Loss and DSD training}
DRN is extended to cope with a dual class prediction: the state of the glass envelope, and the plant where the HCE is located. The DRN is trained for 200 epochs with the same parameters as mentioned before, but using $L_{dual}$ as loss function. $\alpha=0.5$ is maintained for $L_g$, and $\beta_1=\beta_2=1$ gives a similar importance to both losses during the training phase. In addition, DSD training is implemented, both for DRN with $L_g$ loss, and DRN with dual loss. DSD training is performed for a total of 100 epochs, divided in 40 epochs for the learning the important connections, 20 epochs for reinforcing the important connections, and 40 epochs for fine-tuning the network. The selected sparsity was $s=0.8$, representing that the $80\%$ of the weights of the network were used during the Sparse phase.

Table \ref{tab:tab_improved_clfs_test} depicted the results obtained with the aforementioned techniques in the test set. The voting scheme for RF achieve a large improvement in terms of recall, but also a high decrease on the F1-score. This result may be due to the use of the same broken samples among different forests. This ensure better detection of new broken pipes, but misclassifies non-broken HCEs with features similar to a broken sample. On the other hand, both dual loss and DSD training show an improvement in terms of recall, with an acceptable decrease in the F1-score. Specifically, the combination of both techniques in $\text{DRN}_{L_{dual} + DSD}$ boost the performance of the baseline DRN by 5\% in terms of Recall while keeping the same F1-Score. On the other hand, Random Under Sampling increases all the metrics considered for the baseline ML classifiers. The highest improvement is obtained with B-HGBC, resulting in the best F1-score and M-AVG F1 of all the algorithms evaluated. Additionally it worth mentioning the increase of 8\% in the Recall for this technique.

\begin{table}
    \centering
    \begin{tabular}{|c|c|c|c|}
    \hline
    Algorithm & %Precision & 
    Recall & F1 Score & M-AVG F1 \\
    \hline
        %Voting RFs (probs)     &  49 & 85 & 62 & 81 \\
    V-RF     %&  50 
    & 85 (+15) & 63 (-9) & 81 \\
    $\text{DRN}_{L_g+DSD}$     %&  59 
    & 81 (+3) & 68 (-2) & 84  \\
    $\text{DRN}_{L_{dual}}$     %&  59 
    & 80 (+2) & 68 (-2) & 83  \\
    $\text{DRN}_{L_{dual} + DSD}$     %&  57 
    & 83 (+5) & 70 (+0) & 84  \\
    $\textnormal{RF}_{RUS}$     %&  50 
    & 78 (+5) & 73 (+1) & 86 \\
    $\textnormal{B-HGBC}_{RUS}$
    & 74 (+8) & 75 (+3) & 87 \\
    \hline
    \end{tabular}
    \caption{Results for the test set, using a voting scheme for RF and dual loss for DRN. The metrics for DRN shown the value obtained for the broken glass problem. The increase for each metrics with respect to Table \ref{tab:tab_standard_test} is shown in parenthesis.}
    \label{tab:tab_improved_clfs_test}
\end{table}

\section{Ablation Study}
\label{sec:Ablation}

In this section we perform an ablation study using the above techniques to detect the performance boundaries of our system.
Artificial neural networks contains several parameters, turning the design of the architecture in a complex task. Therefore, we conduct an ablation study for the case of the DRN, comparing the impact of different parameters involved in the training process of the network. Specifically, we perform the study for $\text{DRN}_{L_{dual} + DSD}$. 

As depicted in Table \ref{tab:ablation}, changing the model parameters can have a positive or a negative impact on the metrics evaluated. This study helps to define the best values for a given configuration. For instance, the boundaries of $\alpha$ seems to be enclosed by 0 and 1, as for this values the recall and the F1-score, respectively, are highly reduced when compared to the results of the previous experiment. On the other hand, when using Sigmoid as activation function, the recall of the model drops drastically, while keeping a similar F1-score. This behaviour is perhaps related to Sigmoid being over sensitive to class imbalance; therefore, a higher $\alpha$ value should be used when applying this function. Another important remark is the benefits from extending the DSD training phase, and using a smaller value for $\beta_2$ in order to obtain minor improvements in the F1-score and Recall, respectively.

\begin{table}
    \centering
    \begin{tabular}{|c|c|c|c|c|}
    \hline
    \multirow{2}{*}{DSD epochs} & 20-40-20 & 50-10-150 & \multicolumn{2}{c|}{200-50-75}  \\ \cline{2-5}
    &  -1 +1 & +0 +1 & \multicolumn{2}{c|}{+0 +2}  \\ \hline \hline
    
    \multirow{2}{*}{$L_g$ ($\alpha$)} & 0 & 0.25 & 0.75 & 1  \\ \cline{2-5}
      & -13 +5 &  -4 +5 & +3 -9 & +6 -23 \\ \hline \hline
    
    \multirow{2}{*}{DRN activation} & Sigmoid & Tanh & PReLU & ELU  \\ \cline{2-5}
      & -21 +0 &  -4 +1 & -2 +0  & -4 +2 \\ \hline \hline
      
    \multirow{2}{*}{$L_{dual}$ ($\beta_1$, $\beta_2$)} & \multicolumn{2}{c|}{(1, 0.5)} & \multicolumn{2}{c|}{(0.5, 1)}    \\ \cline{2-5}
      & \multicolumn{2}{c|}{+1 +0} &  \multicolumn{2}{c|}{-2 +3}  \\ \hline \hline
      
    \multirow{2}{*}{DSD sparsity} & 0.3 & 0.5 & \multicolumn{2}{c|}{0.9}  \\ \cline{2-5}
      & -2 +1 &  +0 +1 & \multicolumn{2}{c|}{+0 +0} \\ \hline
    \end{tabular}
    \caption{+x,+y represents an increase of x,y, respectively, in the recall and f1 values. The reference is taken from the values of $\text{DRN}_{L_{dual} + DSD}$ in Table \ref{tab:tab_improved_clfs_test}.}
    \label{tab:ablation}
\end{table}

The results obtained with the proposed DRN for the dual problem is depicted on Table \ref{tab:dual_results}. The perfect performance on the test evidences the learning capabilities of our system. In addition, it is an empirical proof that our simulated data provides essential clues about the different processes occurring within each CSP plant. This is in important aspect to remark, as it sustains that both our data and our learning systems can produce valuable information.

\begin{table}
    \centering
    \begin{tabular}{|c|c|c|c|}
    \hline
    Algorithm & Recall & F1 Score & M-AVG F1 \\
    \hline
    $\text{DRN}_{L_{dual}}$      & 100 & 100 & 100  \\
    $\text{DRN}_{L_{dual} + DSD}$      & 100 & 100 & 100  \\
    \hline
    \end{tabular}
    \caption{Results for the test set, considering the dual problem.}
    \label{tab:dual_results}
\end{table}

\section{Conclusion}
\label{sec:conclusion}

Fault detection in CSP systems is mandatory in order to ensure reliability and safety for plants in operation. To address this problem, this paper proposes the first public available dataset, ATSet, for detecting broken glass envelopes of absorber tubes in CSP plants. The dataset contains a large amount of numerical data obtained from seven real plants, and presents a high imbalance in the number of samples of the proposed classes. We study the impact of several descriptive variables extracted from the data set, and perform a correlation analysis to select the most independent features. We observe the advantages of applying Feature Engineering, especially for the dual problem: classifying the HCE depending on the plant where it is located. We provide different tests under different configurations of our data set, using Random Forest, Histogram Gradient Boost Classifiers, and complex Deep Learning techniques. To the best of our knowledge, there is no prior automated solution to this problem using data from operating plants.

Our experiments evidences a boost in performance when: 1) random under-sampling is applied to traditionally Machine Learning methods (Recall increased up to 8\%); 2) a dual optimization with DSD training is applied to the Deep Residual Network (Recall increase up to 5\%). Additionally, we emphasize the learning capabilities of our system and the valuable information contained in our data, by achieving a perfect score in the testing set for the dual problem. 

Visual inspection with UAVs is a promising technique for detecting failures in CSP plants. Adding images to our data set is an interesting line of future research that could improve the performance of the proposed algorithms, by merging their learning capabilities with Convolutional Neural Networks. Finally, other aspects to be addressed in future works is the use of different techniques to overcome class imbalance, such as the creation of virtual data using Generative Adversarial Networks \cite{goodfellow2014generative}.

\section*{Declarations}

\subsection{Funding}

This work is partially supported by the European Union’s Horizon 2020 research and innovation program under the Marie Sklodowska-Curie grant agreement 734922, the Spanish Ministry of Economy and Competitiveness (MTM2016-76272-R AEI/FEDER,UE), the Spanish Ministry of Science and Innovation CIN/AEI/10.13039/501100011033/
(PID2020-114154RB-I00) and the European Union NextGenerationEU/PRTR (DIN2020-011317).

\subsection{Conflict of Interest}

The authors have no conflicts of interest to declare that are relevant to the content of this article.

\subsection{Data availability}
\label{sec:availability}

The data that support the findings of this study will be made available upon reasonable request for academic use. Every request will be reviewed by VirtualMech, and the researcher will need to sign a data access agreement with VirtualMech after approval.

% \subsection{Author Contributions}

% All authors contributed to the study conception and design. Material preparation, data collection and analysis were performed by M.A. Pérez-Cutiño, F. Rodríguez and L.D. Pascual. The first draft of the manuscript was written by J.M. Díaz-Bañez and all authors commented on previous versions of the manuscript. All authors read and approved the final manuscript. Conceptualization and supervision was mainly performed by J.M. Díaz-Bañez.

\subsection{Ethics approval}

All of the authors confirm that there is no potential acts of misconduct in this work, and approve of the journal upholding the integrity of the scientific record.

\subsection{Consent to participate}

The authors consent to participate.

\subsection{Consent for publication}

The authors consent to publish.

% \begin{table}
%     \centering
%     \begin{tabular}{|c|c|c|c|c|}
%     \hline
%      & SCE 01 & SCE 12 & Others & Not measured \\
%     \hline
%     Percent &  41.4 & 32.9 & 15.7 & 10 \\
%     \hline
%     \end{tabular}
% \end{table}

% or


% \appendices

% % you can choose not to have a title for an appendix
% % if you want by leaving the argument blank
% \section{}

% \section*{Acknowledgment}


\ifCLASSOPTIONcaptionsoff
  \newpage
\fi



\bibliographystyle{IEEEtran}
\bibliography{bibtex/bib/IEEEexample.bib}
\end{document}